\documentclass[conference]{IEEEtran}
\IEEEoverridecommandlockouts
\usepackage{cite}
\usepackage{amsmath,amssymb,amsfonts}
\usepackage{algorithmic}
\usepackage{graphicx}
\usepackage{textcomp}
\usepackage{xcolor}
\usepackage{bm}
\usepackage{subcaption}
\usepackage{url}

\newcommand{\etal}{\textit{et al. }}
\newcommand{\trsp}{{\scriptscriptstyle\top}}

\def\BibTeX{{\rm B\kern-.05em{\sc i\kern-.025em b}\kern-.08em
    T\kern-.1667em\lower.7ex\hbox{E}\kern-.125emX}}
\begin{document}

\title{Learning Dynamical System for Grasping Motion\\
\thanks{This work was supported by National Key R\&D Program of China (2018YFB2100903). \emph{(Corresponding author: Xiaohui Xiao.)}}
}

\author{Xiao Gao, Miao Li and Xiaohui Xiao% <-this % stops a space
\thanks{Hubei Key Laboratory of Waterjet Theory and New Technology, Wuhan University, 430072 Wuhan, China (xiaogao@whu.edu.cn, miao.li@whu.edu.cn, xhxiao@whu.edu.cn).}%
}%
% \author{\IEEEauthorblockN{1\textsuperscript{st} Given Name Surname}
% \IEEEauthorblockA{\textit{dept. name of organization (of Aff.)} \\
% \textit{name of organization (of Aff.)}\\
% City, Country \\
% email address or ORCID}
% \and
% \IEEEauthorblockN{2\textsuperscript{nd} Given Name Surname}
% \IEEEauthorblockA{\textit{dept. name of organization (of Aff.)} \\
% \textit{name of organization (of Aff.)}\\
% City, Country \\
% email address or ORCID}
% \and
% \IEEEauthorblockN{3\textsuperscript{rd} Given Name Surname}
% \IEEEauthorblockA{\textit{dept. name of organization (of Aff.)} \\
% \textit{name of organization (of Aff.)}\\
% City, Country \\
% email address or ORCID}
% \and
% \IEEEauthorblockN{4\textsuperscript{th} Given Name Surname}
% \IEEEauthorblockA{\textit{dept. name of organization (of Aff.)} \\
% \textit{name of organization (of Aff.)}\\
% City, Country \\
% email address or ORCID}
% \and
% \IEEEauthorblockN{5\textsuperscript{th} Given Name Surname}
% \IEEEauthorblockA{\textit{dept. name of organization (of Aff.)} \\
% \textit{name of organization (of Aff.)}\\
% City, Country \\
% email address or ORCID}
% \and
% \IEEEauthorblockN{6\textsuperscript{th} Given Name Surname}
% \IEEEauthorblockA{\textit{dept. name of organization (of Aff.)} \\
% \textit{name of organization (of Aff.)}\\
% City, Country \\
% email address or ORCID}
% }

\maketitle

\begin{abstract}
    Dynamical System has been widely used for encoding trajectories from human demonstration, which has the inherent adaptability to dynamically changing environments and robustness to perturbations. In this paper we propose a framework to learn a dynamical system that couples position and orientation based on a  diffeomorphism. Different from other methods, it can realise the synchronization between positon and orientation during the whole trajectory. Online grasping experiments are carried out to prove its effectiveness and online adaptability.
\end{abstract}


% !TEX root = ../root.tex

\section{Introduction}
\label{sec::intro}
% \cite{cho2005stable}. 

%  imitation learning, dynamical system
Generating robot motion is one of the most important parts in robotics. Imitation learning\cite{Aude2020review} can be adopted to encode human demonstrations and tranfer skills to robots, which offer a convenient way for robot motion planning. In this paper, we focus on the time-invariant dynamical system , which can be robust to temporal and spatial perturbation and can adapt to dynamically changing environments.

% time-invariant DS
% seds and other DS method
Many dynamical systems for reaching movements have been proposed based on imitation learning, which encode human demonstration trajectories as a nonlinear system $\dot{\bm x} = \bm f(\bm x)$, where $\bm x$ is the robot state and $\dot{\bm x}$ is its velocity. In order to make sure any initial point in the system can convergence to the only attractor (equilibrium point), Khansari \etal \cite{khansari2011SEDS} proposed the Stable Estimator of Dynamical System (SEDS) algorithm to learn a global asymptotically stable DS. They used a quadratic Lyapunov function as constraints when learning SEDS by Gaussian mixture models. Due to the Lyapunov function, SEDS can only generate contractive motion. To overcome this problem and represent more complex motion, many different methods were proposed \cite{khansari2014SEDS,lemme2014neural,  neumann2015tauSEDS, perrin2016fastDM, Urain2020iflow}.

% problem, it cannot make sure the synchronization between position and orientation
However, when it comes to Cartesian (task) space motion planning, the above methods only considered position information and ignored orientation modeling. The orientation of the robot end-effector was constant in most DS papers, or it was generated by interpolation, which means that the position and orientation are not synchronous.

% For example, in \cite{Gribovskaya}, axis/angle representation was used for orientation representation and two functions were learnt for the position DS and orientation DS seperately. Kim \etal \cite{Kim2014catching} used SEDS for the end-effector position and the orientation was generated by axis/angle interpolation. When we hope to find a skill model that produces both position and orientation trajectories similar to the demonstrated ones, a coupled DS is required, since an alone orientation planner is meaningful only if it is at the proper position.

% conclude our work, 
% our contribution
 In our previous paper \cite{Gao2021motion}, a diffeomorphism\footnote{A diffeomorphism is an invertible and continuously differentiable function} was proposed to mapping the pose between two spaces. Inspired by Perrin's work \cite{perrin2016fastDM}, in this paper, we propose a coupled DS to realise the synchronization between position and orientation. Two diffeomorphisms are applied on complex demonstration trajectories to receive simple trajectories in latent spaces. Then a global asymptotically stable DS is designed to encode pose trajectories in one latent space, which is mapped back to the original demonstration for generating the required DS. Our main contribution is as follows:
 \begin{enumerate}
     \item a framework for the coupled DS with synchronization between position and orientation;
     \item guarantee of the global asymptotically stability for the coupled DS;
     \item inherent robustness and adaptability for both position and orientation.
 \end{enumerate}

% The method of the coupled DS is described in Section \ref{sec::method}. And grasping experiments under perturbations and dynamically changing environments are showed to prove the ability of our method in Section \ref{sec::exp}.

% !TEX root = ../root.tex

\section{Method}
\label{sec::method}

\begin{figure}[t]
    \centering
    \includegraphics[width=0.5\textwidth]{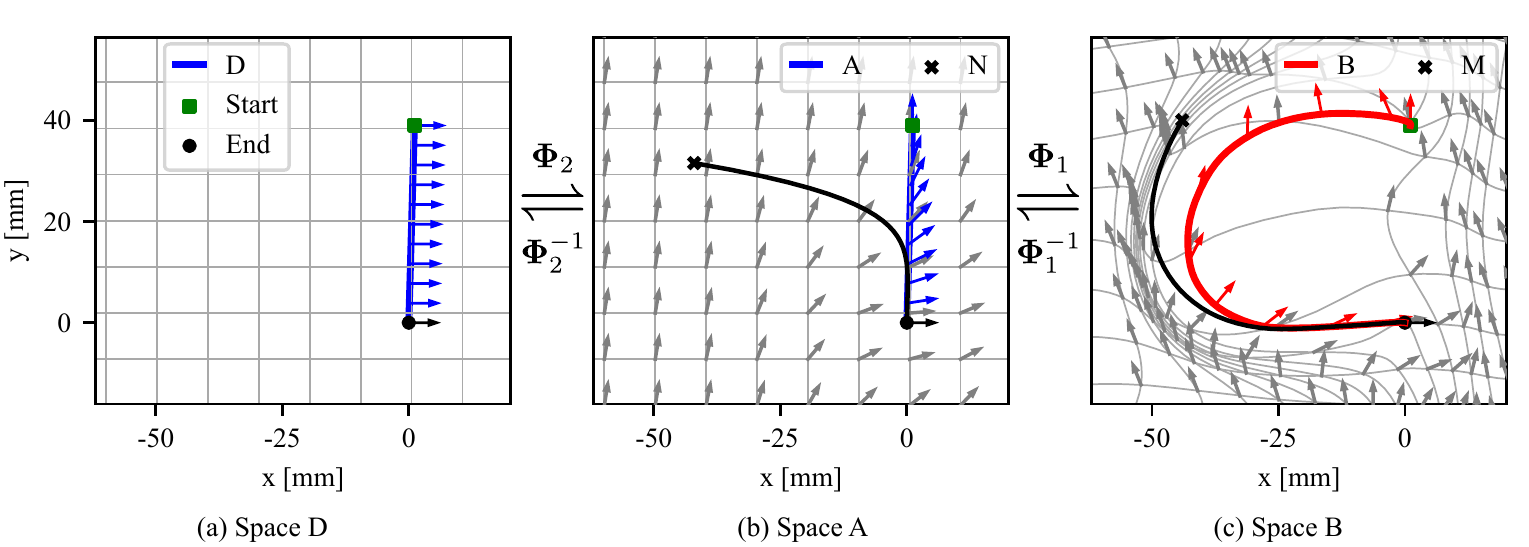}
    \caption{Two diffeomorphisms for pose mappings   }
    \label{fig-DAB}
  \end{figure}
%  demonstration trajectory and its transform, DAB figures,  two diffeomorphisms

Assuming that there is a human demonstration trajectory $\bm B = \{\bm x'_{i}, \bm q'_{i}\}_{i=1}^N $, which is plotted in Fig.~\ref{fig-DAB}(c) as red the curve for position and red arrows (x-axis direction) for orientation. Here we use quaternion representation. The end pose of trajectory $\bm B$ is as ${\bm 0, \bm q_I}$, also as the attractor. $\bm q_I = [1,0,0,0]^\trsp$ is the identity quaternion.
A simple trajectory $\bm A$ in space A (Fig.~\ref{fig-DAB}(b)) is generated by a linear interpolation of the start point of $\bm B$ as:
\begin{equation}
    \bm A = \{\bm x_{i}, \bm q_{i}\}_{i=1}^N = \{ t \bm x_{1}' ,{\bm q'_{1}}^t \}, t = \frac{N-i}{N-1} (i = 1,2,...,N)  ,
    \label{eq-B2A}
  \end{equation}
Blue arrows in Fig.~\ref{fig-DAB}(b) show the orientation of $\bm A$. In Fig.~\ref{fig-DAB}(c), trajectory $\bm D =\{ \bm x_i, \bm q_I\}_{i=1}^N$, where the position is the same as $\bm A$ and orientation is constant at identity quaternion.

Based on our previous diffeomorphic mapping framework for motion mapping \cite{Gao2021motion}, and set an arbitrary pose $\bm N=\{\bm x,\bm q\}$ in space A, we build a diffeomorphism $\bm \Phi_1$ between the space A and B to map the trajectory $\bm A$ to $\bm B$, and another diffeomorphism $\bm \Phi_2$  between the space D and A to map the trajectory $\bm D$ to $\bm A$. The two diffeomorphisms are as follows:
\begin{equation}
    \bm \Phi_1:\left\{
    \begin{aligned}
      \bm x' &= \bm h_{1}(\bm x)    ,\\
       \bm q'& = \bm g_1(\bm x) * \bm q ,
    \end{aligned}
    \right.
    \label{3eq-A2B}   
    \bm \Phi_2: \left\{
      \begin{aligned}
        \bm x &= \bm h_{2}(\bm x_D) = \bm x_D , \\
         \bm q& = \bm g_2(\bm x_D) * \bm q_I = \bm g_2(\bm x) ,
      \end{aligned}
      \right.
    %   \label{3eq-D2A}
  \end{equation}
where $(\bm x_D, \bm q_I)$ is the corresponding pose of $\bm \Phi_2^{-1}(\bm x, \bm q)$, and $\bm g_1(\bm x) ,\bm g_2(\bm x) \in \mathcal{S}^3$ are unit quaternions, and also functions of position $\bm x$.

To visualize the two diffeomorphisms, we set equal-spaced grid points in space D (gray lines in Fig.~\ref{fig-DAB}(a)), with identify quaternions. Under the two diffeomorphisms $\bm \Phi_1$ and $\bm \Phi_2$, grid points are distorted in the space A and B, where when grid points are close to trajectories $\bm A$ or $\bm B$, the orientations of the points are also close to the corresponding orientation direction (Fig.~\ref{fig-DAB}(b)--(c)).

% DS
The position DS in space A is as:
\begin{equation}
    \dot{\bm x} = \bm f_1(\bm x) = \gamma_1(\bm x) \bm P \bm x, \gamma(\bm x)>0 .
\label{eq-A1}
\end{equation}
where the ymmetric negative definite matrix $\bm P$ is designed based on $\bm A$. $\gamma_1(\bm x)$ is to adjust the velocity.
For orientation, we set the angular velocity as:
\begin{subequations}
    \begin{align}
      \bm \omega &=  \gamma_2(\bm q)\beta \log(\bm q *  \bar{\bm g}_2(\bm x) ) + \bm \omega_r
      ,\gamma_2(\bm q)>0, \beta<0\label{eq-A2}\\
      \bm \omega_r &= - 2\bm q * \frac{\partial \bar{\bm g}_2(\bm x)   }{\partial \bm x}\dot{\bm x} * \bm g_2(\bm x) * \bar{\bm q} 
      \label{3eq-A3}
    \end{align}
    \label{eq-A}
  \end{subequations}
which includes a feedback term and a feedforward term. The goal is to track the desired orientation $\bm g_2(\bm x)$. So, the angular velocity $\bm \omega$ is a function of current position $\bm x$ and orientation $\bm q$. The stability can be proved by the quadratic Lyapunov functions for both position and orientation. Finally, we can use the diffeomorphism $\bm \Phi_1$ to compute the coupled DS in space B.

% !TEX root = ../root.tex

\vspace{-6pt}
\section{Experiments}
\label{sec::exp}
\begin{figure}[t]
    \centering
   \begin{subfigure}[b]{0.11\textwidth}
       \centering
       \includegraphics[width=\textwidth, trim=300 610 0 1100 ,clip]{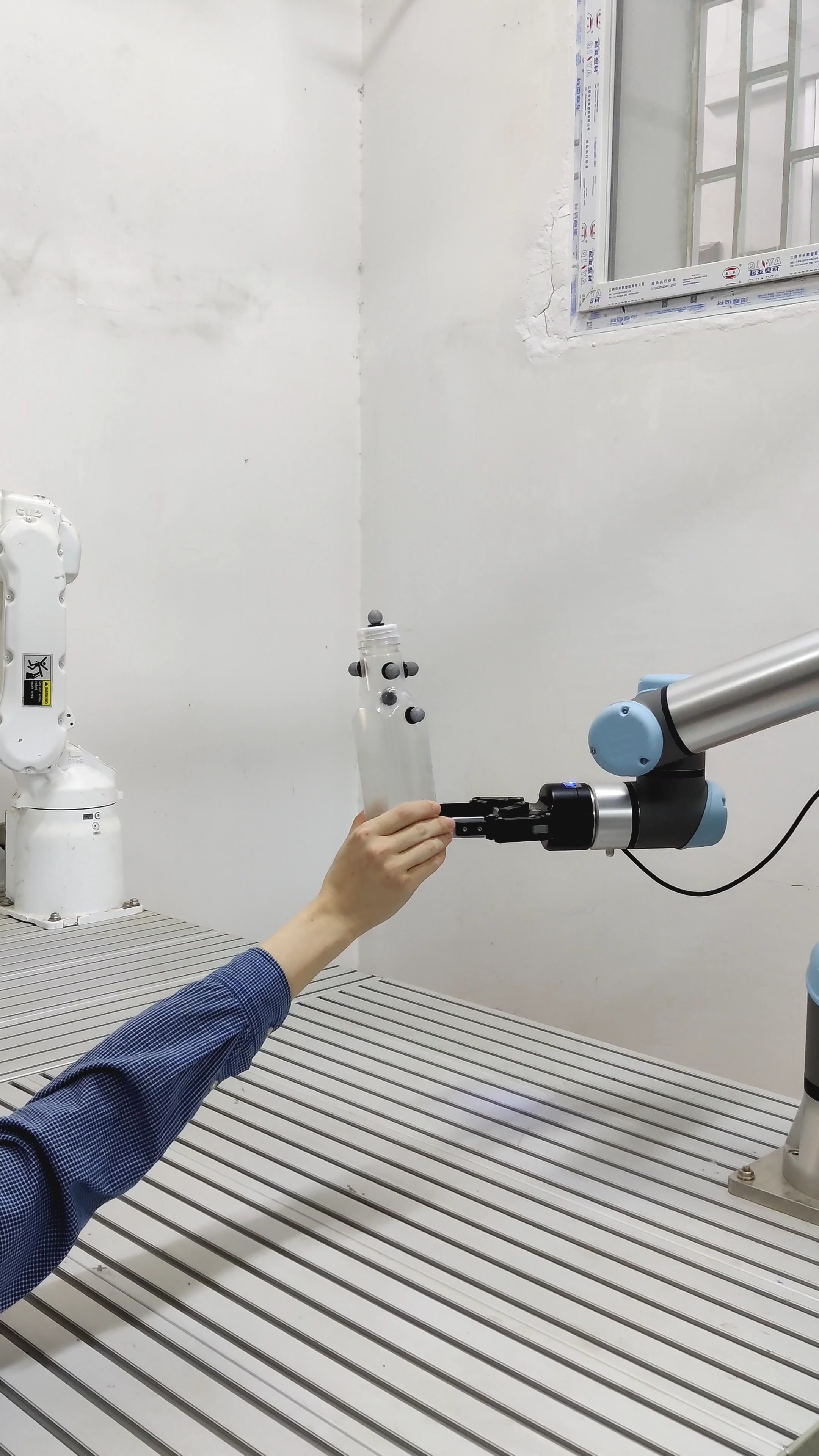}  
       \caption{}
         \label{3fig-s1}
   \end{subfigure}
%    \hspace{2pt}
    % \begin{subfigure}[b]{0.23\textwidth}
    %       \centering 
    %       \includegraphics[width=\textwidth, trim=300 610 0 1100,clip]{figures/CH3/1/s2.jpg}  
    %       \caption{}  %左、下、右、上
    %       \label{3fig-s2}
    %       \end{subfigure} \hspace{2pt}
    % \begin{subfigure}[b]{0.23\textwidth}
    %   \centering 
    %   \includegraphics[width=\textwidth, trim=300 610 0 1100,clip]{figures/CH3/1/s3.jpg}   
    %   \caption{}
    %   \label{3fig-s3}
    % \end{subfigure} \hspace{2pt}
    \begin{subfigure}[b]{0.11\textwidth}
        \centering 
        \includegraphics[width=\textwidth, trim=300 610 0 1100,clip]{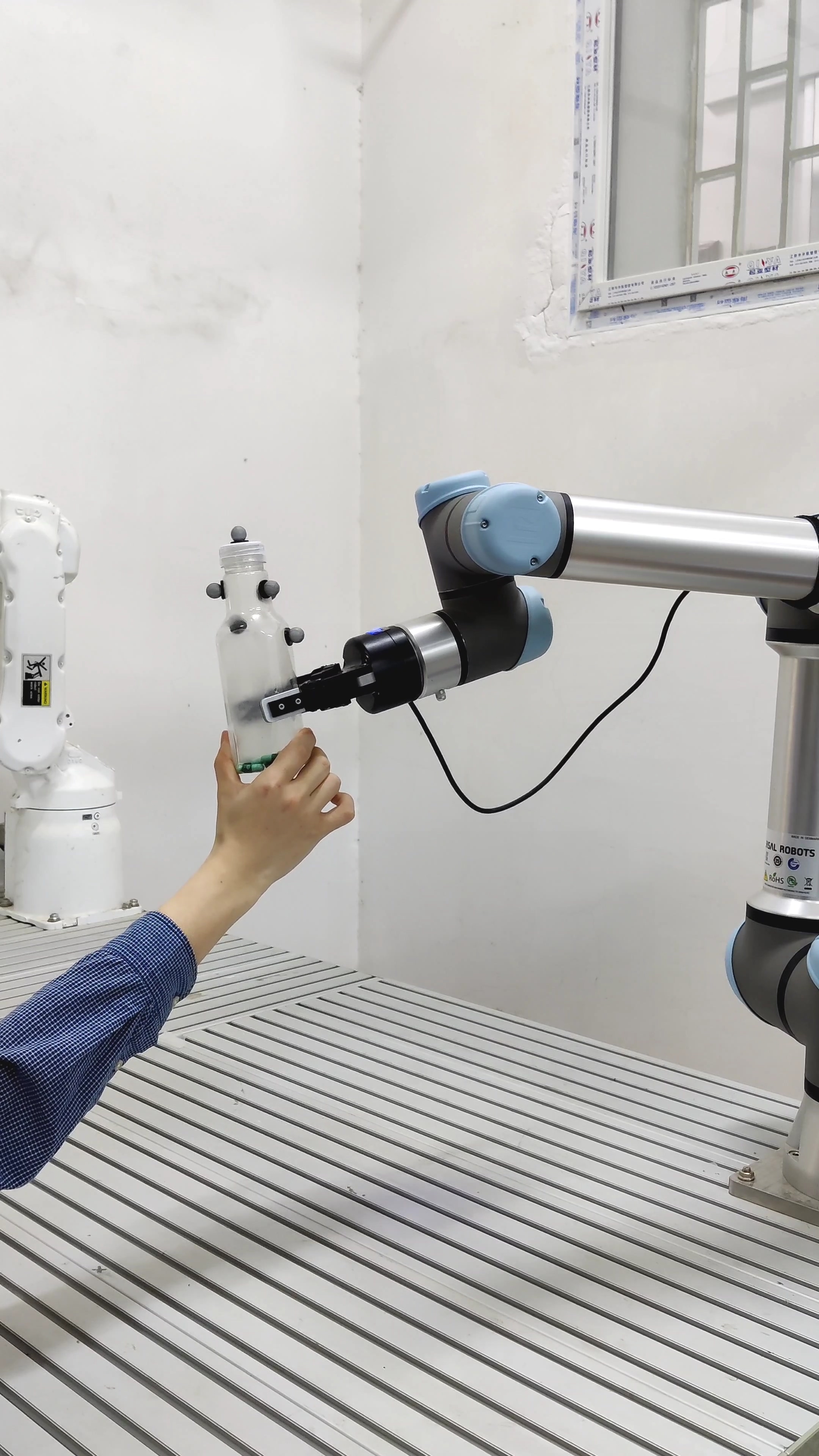}   
        \caption{}
        \label{3fig-s4}
        \end{subfigure}   
    %     \begin{subfigure}[b]{0.23\textwidth}
    %       \centering
    %       \includegraphics[width=\textwidth, trim=300 610 0 1100,clip]{figures/CH3/1/s5.jpg}  
    %       \caption{}
    %         \label{3fig-s5}
    %   \end{subfigure}
    %   \hspace{2pt}
       \begin{subfigure}[b]{0.11\textwidth}
             \centering 
             \includegraphics[width=\textwidth, trim=300 810 0 900,clip]{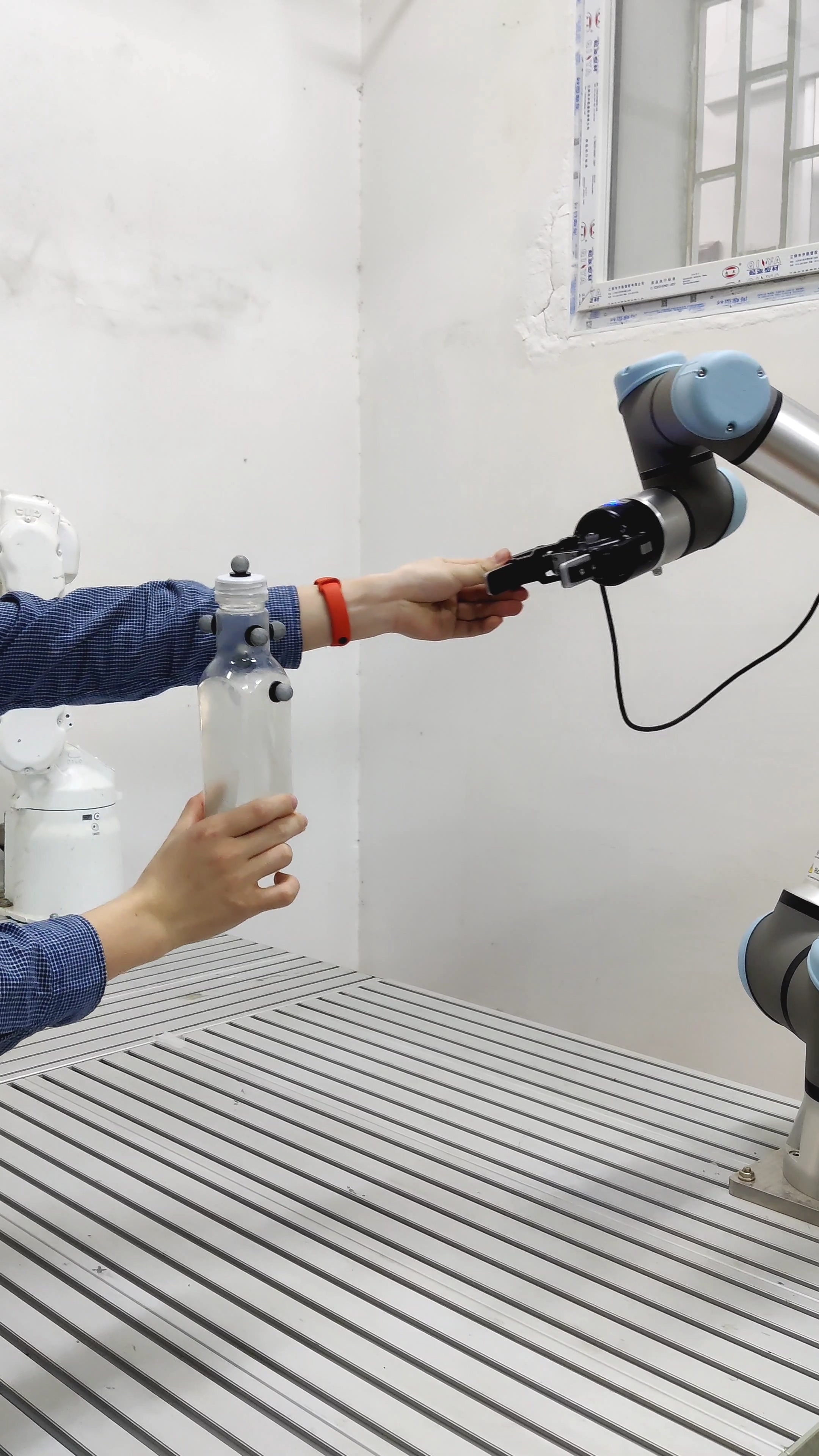}  
             \caption{}  %左、下、右、上
             \label{3fig-s6}
             \end{subfigure} 
            %  \hspace{2pt}
    %    \begin{subfigure}[b]{0.23\textwidth}
    %      \centering 
    %      \includegraphics[width=\textwidth,  trim=300 810 0 900,clip]{figures/CH3/1/s7.jpg}   
    %      \caption{}
    %      \label{3fig-s7}
    %    \end{subfigure} \hspace{2pt}
       \begin{subfigure}[b]{0.11\textwidth}
           \centering 
           \includegraphics[width=\textwidth,  trim=300 810 0 900,clip]{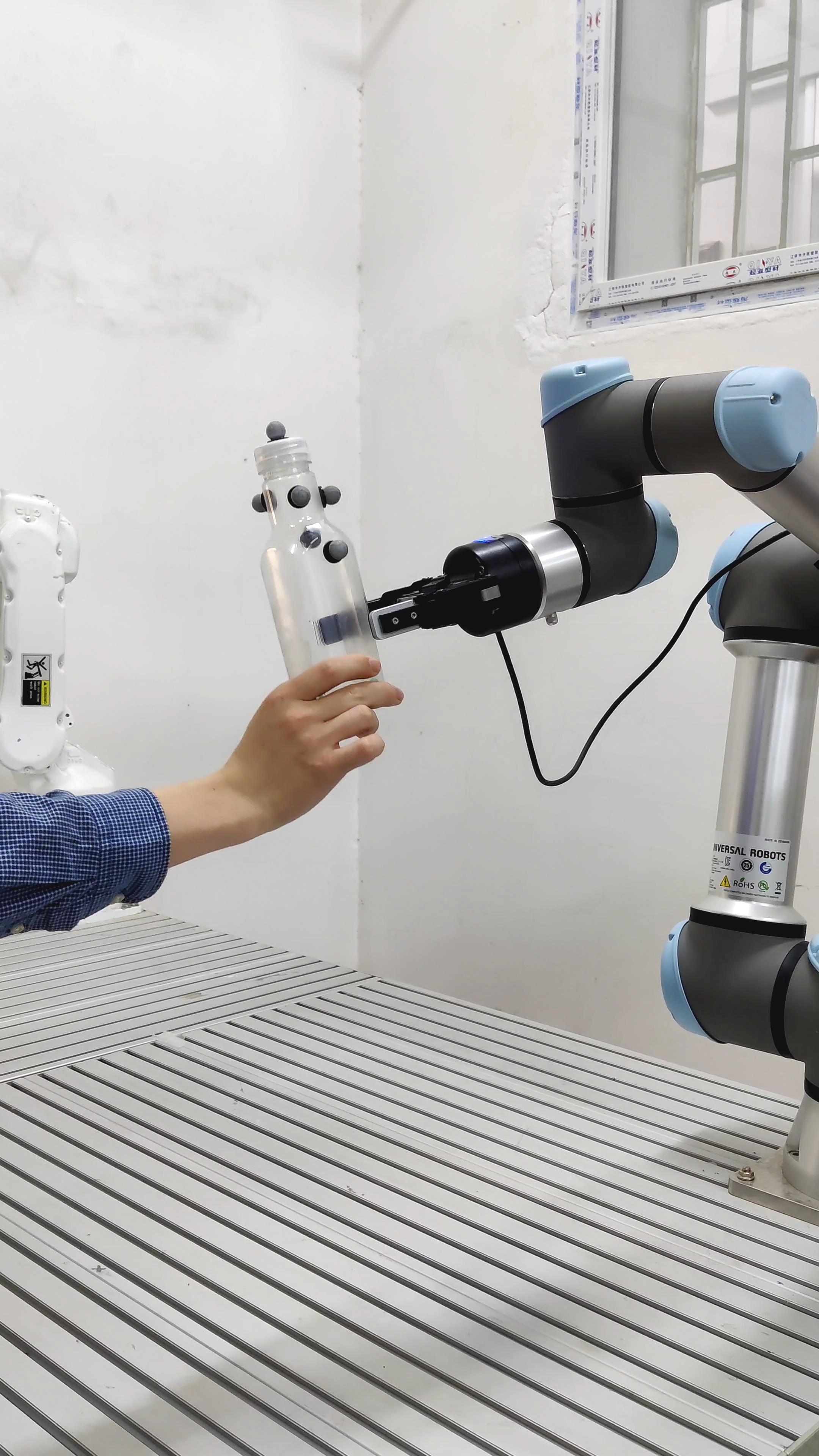}   
           \caption{}
           \label{3fig-s8}
           \end{subfigure} 
  \caption{Snapshots of dynamical grasping}
  \label{fig-snap}
  \end{figure}

Experimental setup is shown in Fig.~\ref{fig-snap}. Six markers of Vicon motion tracking system are attached on a bottle. The goal is to grasp it with the UR5e robot and the Robotiq 85 gripper. The coupled DS is generated by the method in Section \ref{sec::method}.

\begin{figure}[t]
    \centering
    \includegraphics[scale=1]{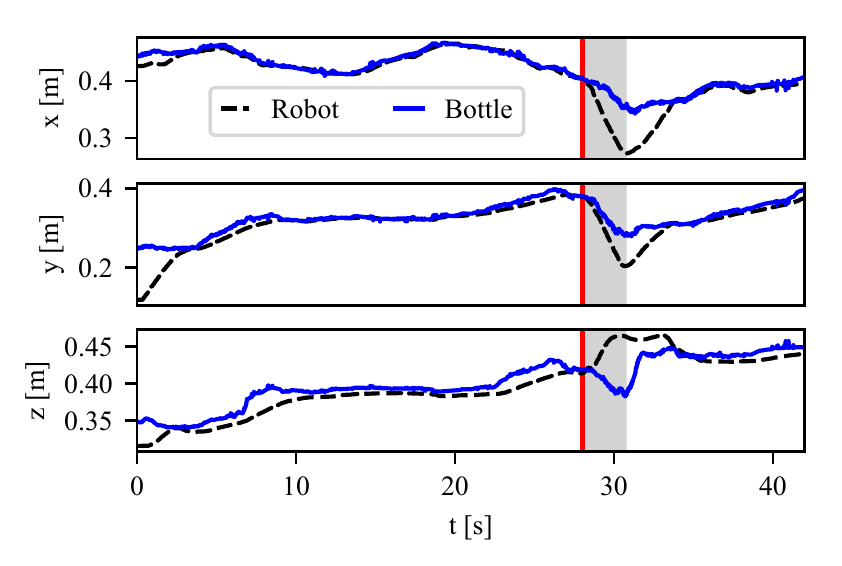}
    \vspace{-8pt}
    \caption{Position of the robot and bottle }
    \label{fig-exp-pos}
  \end{figure}

\begin{figure}[t]
    \vspace{-8pt}
    \centering
    \includegraphics[scale=1]{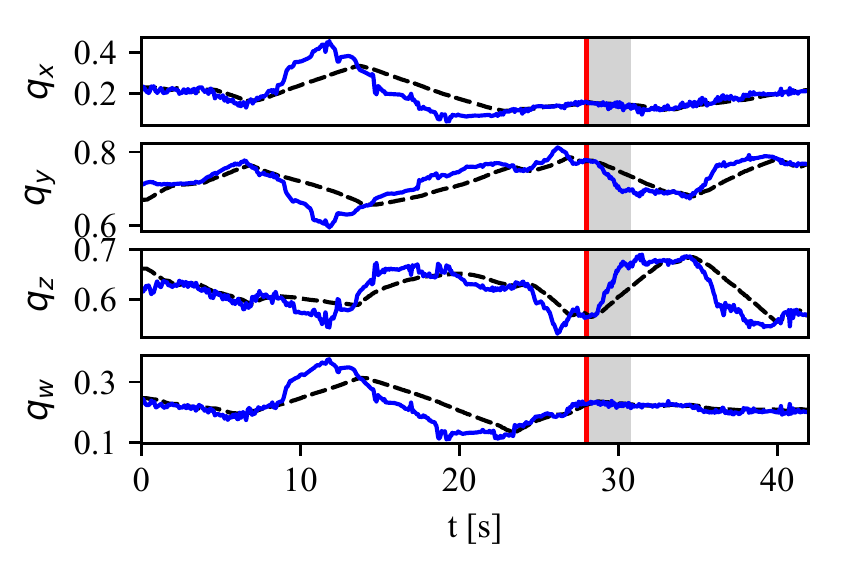}
    \vspace{-8pt}
    \caption{Orientations of the robot and bottle}
    \label{fig-exp-ori}
  \end{figure}

Fig.~\ref{fig-snap} shows the snapshots of the experiments\footnote{The video can be found here \url{https://youtu.be/-V4i8vManVQ}.}. The robot/bottle positon and orientation are plotted in Fig. \ref{fig-exp-pos} and \ref{fig-exp-ori}. We can see that the robot pose was tracking the pose of the bottle, while the pose of bottle was changing due to the movements of the user's right hand. When $t=$ 28-31 s (Fig.~\ref{fig-snap}(c)), a disturbance was applied on the robot by the user's left hand. The robot can also moved to the desired grasping pose.

% !TEX root = ../root.tex

\vspace{-12pt}
\section{Conclusion}
\label{sec::conclusion}

In this paper, we propose a framework of  coupled DS for the generation of linear velocity and angular velocity synchronously. The dynamical grasping experiments show that it can adapt to dynamical environments and can be robust to perturbations.

\bibliographystyle{IEEEtran}	
\bibliography{refer.bib}

% Generated by IEEEtran.bst, version: 1.12 (2007/01/11)
\begin{thebibliography}{1}
\providecommand{\url}[1]{#1}
\csname url@samestyle\endcsname
\providecommand{\newblock}{\relax}
\providecommand{\bibinfo}[2]{#2}
\providecommand{\BIBentrySTDinterwordspacing}{\spaceskip=0pt\relax}
\providecommand{\BIBentryALTinterwordstretchfactor}{4}
\providecommand{\BIBentryALTinterwordspacing}{\spaceskip=\fontdimen2\font plus
\BIBentryALTinterwordstretchfactor\fontdimen3\font minus
  \fontdimen4\font\relax}
\providecommand{\BIBforeignlanguage}[2]{{%
\expandafter\ifx\csname l@#1\endcsname\relax
\typeout{** WARNING: IEEEtran.bst: No hyphenation pattern has been}%
\typeout{** loaded for the language `#1'. Using the pattern for}%
\typeout{** the default language instead.}%
\else
\language=\csname l@#1\endcsname
\fi
#2}}
\providecommand{\BIBdecl}{\relax}
\BIBdecl

\bibitem{Aude2020review}
H.~Ravichandar, A.~S. Polydoros, S.~Chernova, and A.~Billard, ``{Recent
  Advances in Robot Learning from Demonstration},'' \emph{Annual Review of
  Control, Robotics, and Autonomous Systems}, vol.~3, no.~1, pp. 297--330,
  2020.

\bibitem{khansari2011SEDS}
S.~M. Khansari-Zadeh and A.~Billard, ``Learning stable nonlinear dynamical
  systems with gaussian mixture models,'' \emph{IEEE Transactions on Robotics},
  vol.~27, no.~5, pp. 943--957, 2011.

\bibitem{khansari2014SEDS}
------, ``Learning control {L}yapunov function to ensure stability of dynamical
  system-based robot reaching motions,'' \emph{Robotics and Autonomous
  Systems}, vol.~62, no.~6, pp. 752--765, 2014.

\bibitem{lemme2014neural}
A.~Lemme, K.~Neumann, R.~F. Reinhart, and J.~J. Steil, ``Neural learning of
  vector fields for encoding stable dynamical systems,'' \emph{Neurocomputing},
  vol. 141, pp. 3--14, 2014.

\bibitem{neumann2015tauSEDS}
K.~Neumann and J.~J. Steil, ``Learning robot motions with stable dynamical
  systems under diffeomorphic transformations,'' \emph{Robotics and Autonomous
  Systems}, vol.~70, pp. 1--15, 2015.

\bibitem{perrin2016fastDM}
N.~Perrin and P.~Schlehuber-Caissier, ``Fast diffeomorphic matching to learn
  globally asymptotically stable nonlinear dynamical systems,'' \emph{Systems
  \& Control Letters}, vol.~96, pp. 51--59, 2016.

\bibitem{Urain2020iflow}
J.~{Urain}, M.~{Ginesi}, D.~{Tateo}, and J.~{Peters}, ``Imitationflow: Learning
  deep stable stochastic dynamic systems by normalizing flows,'' in \emph{Proc.
  {IEEE/RSJ} Intl Conf. on Intelligent Robots and Systems ({IROS})}, 2020, pp.
  5231--5237.

\bibitem{Gao2021motion}
X.~Gao, J.~Silv\'erio, E.~Pignat, S.~Calinon, M.~Li, and X.~Xiao, ``Motion
  mappings for continuous bilateral teleoperation,'' \emph{IEEE Robotics and
  Automation Letters ({RA-L})}, vol.~6, no.~3, pp. 5048--5055, 2021.

\end{thebibliography}

\end{document}